\documentclass{article}

\usepackage[final,nonatbib]{nips_2018}

\usepackage[utf8]{inputenc} 
\usepackage[T1]{fontenc}    
\usepackage{hyperref}       
\usepackage{url}            
\usepackage{booktabs}       
\usepackage{amsfonts}       
\usepackage{nicefrac}       
\usepackage{microtype}      

\usepackage{multirow}
\usepackage{graphicx}
\usepackage{adjustbox}
\usepackage{array}
\usepackage{multirow}
\usepackage{xargs}  
\usepackage{wrapfig}

\usepackage[pdftex,dvipsnames]{xcolor}  
\usepackage[colorinlistoftodos,prependcaption,textsize=tiny]{todonotes}
\newcommandx{\unsure}[2][1=]{\todo[linecolor=red,backgroundcolor=red!25,bordercolor=red,#1]{#2}}
\newcommandx{\change}[2][1=]{\todo[linecolor=blue,backgroundcolor=blue!25,bordercolor=blue,#1]{#2}}
\newcommandx{\info}[2][1=]{\todo[linecolor=OliveGreen,backgroundcolor=OliveGreen!25,bordercolor=OliveGreen,#1]{#2}}
\newcommandx{\improvement}[2][1=]{\todo[linecolor=Plum,backgroundcolor=Plum!25,bordercolor=Plum,#1]{#2}}
\newcommandx{\thiswillnotshow}[2][1=]{\todo[disable,#1]{#2}}

\newcolumntype{R}[2]{%
    >{\adjustbox{angle=#1,lap=\width-(#2)}\bgroup}%
    l%
    <{\egroup}%
}

\newcommand*\rot{\multicolumn{1}{R{45}{1em}}}

\usepackage{subcaption}
\usepackage{tikz}
\usepackage{pgfplots}
\usepackage{verbatim}

\usetikzlibrary{pgfplots.groupplots}
\pgfplotsset{compat=1.11}
\usepgfplotslibrary{ternary}
\usepackage{tikz}
\usetikzlibrary{arrows.meta}
\usepackage{tkz-graph}
\usetikzlibrary{positioning,automata}
\usetikzlibrary{calc,spy,shapes}
\usetikzlibrary{arrows,decorations.pathmorphing,fit,positioning,patterns}
\usetikzlibrary{decorations.text}
\usepackage{commath}
\usepackage{booktabs}
\usepackage[moderate]{savetrees}

\pgfplotsset{compat=newest}

\definecolor{lcolor}{HTML}{268bd2}
\definecolor{gcolor}{HTML}{c11b17}
\definecolor{wcolor}{HTML}{859900}
\definecolor{ccolor}{HTML}{ffcc00}

\newcommandx{\squad}{SQuAD}

\usepackage{setspace}
\title{Incremental Reading for Question Answering}

\author{
  Samira Abnar\thanks{Work done during Samira's Internship at Google.} \\
  University of Amsterdam\\
  \texttt{s.abnar@uva.nl} \\
   \And
   Tania Bedrax-Weiss \\
   Google \\
   \texttt{tbedrax@google.com} \\
   \AND
   Tom Kwiatkowski \\
   Google \\
   \texttt{tomkwiat@google.com } \\
   \And
   William W Cohen \\
   Google \\
   \texttt{wcohen@google.com } \\
}

\begin{document}

\maketitle

\begin{abstract}
Any system which performs goal-directed continual learning must not only learn incrementally but process and absorb information incrementally. Such a system also has to understand when its goals have been achieved.  In this paper, we consider these issues in the context of question answering.  Current state-of-the-art question answering models reason over an entire passage, not incrementally.  As we will show, naive approaches to incremental reading, such as restriction to unidirectional language models in the model, perform poorly. We present extensions to the DocQA \cite{clark2017simple} model to allow incremental reading without loss of accuracy.  The model also jointly learns to provide the best answer given the text that is seen so far and predict whether this best-so-far answer is sufficient.
\end{abstract}

\section{Introduction}
Humans can read and comprehend text incrementally. For instance, given a piece of text, our mental state gets updated as we read \cite{traxler1997influence}. We do not necessarily wait until the end of a long document to understand its first sentence. This incremental reading mechanism allows us to avoid consuming more input if we have already reached our goal, and is analogous to the problem faced by a goal-directed continuous learning system, which must also incrementally absorb new information, determine how to use it, and determine if its goals have been achieved.
Inspired by how humans read and learn from text incrementally, we introduce incremental models for text comprehension. 
Our primary goal is to address the problem of incremental reading in the context of language comprehension and Question Answering (QA). 
We formulate the problem as designing a model for question-answering that consumes text incrementally.

By design and definition, Recurrent Neural Networks (RNNs), e.g. Long Short-Term Memory networks (LSTMs) \cite{hochreiter1997long}, process data sequentially, and update their internal states as they read the new tokens. However, on tasks like Question Answering, in all the existing well-performing models, RNNs are employed in a bidirectional way, or a self-attention mechanism is employed \cite{yu2018qanet, clark2017simple, hermann2015teaching, seo2016bidirectional}. This means these models need to processes the whole input sequence to compute the final answer. This is a reasonable approach if the input sequence is as short as a sentence, but it becomes less effective and efficient as the length of the input sequence increases. 

We introduce a new incremental model based on DocQA\cite{clark2017simple}, which is an RNN based model proposed for QA. The incremental DocQA performs similarly to the original system but can process the input text incrementally.
\thiswillnotshow{We propose a mechanism to make this model incremental without losing performance.}
\thiswillnotshow{the number of tokens in each slice should be optimized.}
We propose the use of \emph{slicing} to build incremental models. Slicing RNNs were introduced in \cite{yu2018sliced} with the motivation of enabling parallelization and speeding up sequence processing. Here, we explore using slicing to facilitate incremental processing of the input sequence.

Besides the fact that incremental reading is more plausible from the cognitive perspective, it can also provide an inductive bias for the models, which will make it easier for them to find a more generalizable solution \cite{battaglia2018relational}. 
\thiswillnotshow{We need some examples, for example in sentence comprehension, that incremental decoding of the meaning of the sentence is very useful. Or maybe a completely different task.}
\thiswillnotshow{Our experiments can also reveal some of the properties of the solutions the baseline models find for solving the QA tasks. example: e.g. how long the long-term dependencies are.}
Moreover, incrementality allows the model to be applied to tasks where we do not have the whole input in the beginning, and we need to compute some intermediate outputs. E.g. when the input is a stream, or we are in the context of a dialogue.
We observed that, even if the whole input text is available, it is not always necessary to read the whole text to answer a given question. \thiswillnotshow{either add the experiment or remove!}
In fact, our model achieves its highest performance when it is limited to read only a few tokens after the answer, rather than when it is allowed access to the entire context.
Considering this, we also augment the incremental DocQA with an ``early stopping'' strategy, where it stops consuming the rest of the input text as soon as it is confident that it has found the answer. 
Learning to stop reading or reasoning when the goal is reached is addressed for tasks like text classification \cite{dulac2011text} and question answering \cite{shen2017reasonet,yu2017learning, johansen2017learning}, but the main challenge is that implementing early stopping strategies is only possible when we have an incremental model.





\section{Architecture of the models}
\begin{figure}[h]
\begin{subfigure}[b]{0.5\textwidth}
\centering
\includegraphics[width=\textwidth]{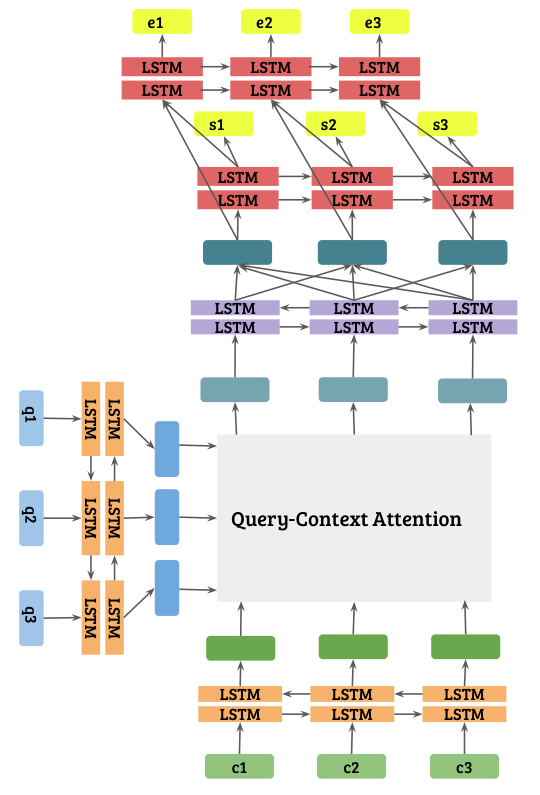}
\caption{\label{fig:docqa}\fontsize{8}{8}\selectfont{DocQA}}
\end{subfigure}
~
\begin{subfigure}[b]{0.5\textwidth}
\centering
  \includegraphics[width=\textwidth]{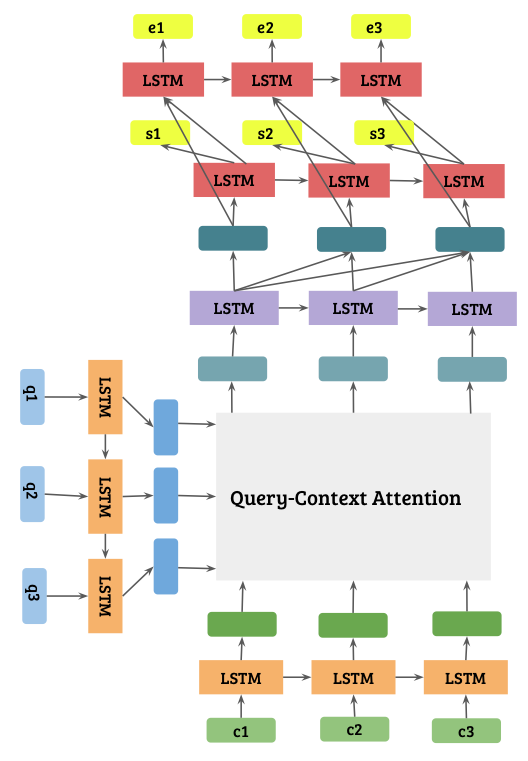}
  \caption{\label{fig:unidocqa}\fontsize{8}{8}\selectfont{Uni-directional DocQA}}
\end{subfigure}
\caption{\fontsize{8}{10}\selectfont{Architecture of Baseline Models. In these figures $q_i$ refers to question tokens, $c_i$ refers to context tokens, and $s_i$ and $e_i$ refer to probability of each token being the start and end of the answer span.}}
\end{figure}

We use DocQA as the baseline model \cite{clark2017simple}.
The architecture of DocQA is illustrated in Figure \ref{fig:docqa}.
\thiswillnotshow
{
The model consists of these layers:
\begin{itemize}
    \item Word embedding layer: Word embedding layer provides the representation for each word in isolation. In DocQA, this layer is a combination of character level and word level word embeddings. 
    \item Contextualized word embedding layer: A recurrent layer, which its purpose is to augment the word representations with contextual information.
    \item Bidirectional attention flow layer:
    The intention of this layer is to compute query aware context representations. It computes the attention weights based on a similarity matrix between question and context terms.
    \item modelling layer: Another RNN layer on top of the bidirectional attention flow layer.
    \item self-attention layer: This is a static attention layer that determines to which steps the model should attend in order to predict the logits for each step. 
    \item prediction layer: Given the attended representations, computes the start and end logits.
\end{itemize}
}
The output of this model is two vectors with the length of the given context. One of the vectors indicates the probability of each token in the context to be the start token of the answer span.
The other vector indicates the probability of each token in the context to be the end of the answer span.
The gold labels indicate the ground truth begin and end of the answer spans. 
\thiswillnotshow{Note that one question might have multiple gold answer spans.}
DocQA is inherently bidirectional thus making it hard to process the input incrementally.  It is possible to remove the bidirectionality by replacing the bidirectional layers with single-directional layers and replacing the global attention with an attention layer that only attends to the past, but this change significantly reduces performance.  
\begin{figure}[t!]
    \centering
    \begin{subfigure}[b]{0.4\textwidth}
        \centering
  \includegraphics[width=\textwidth]{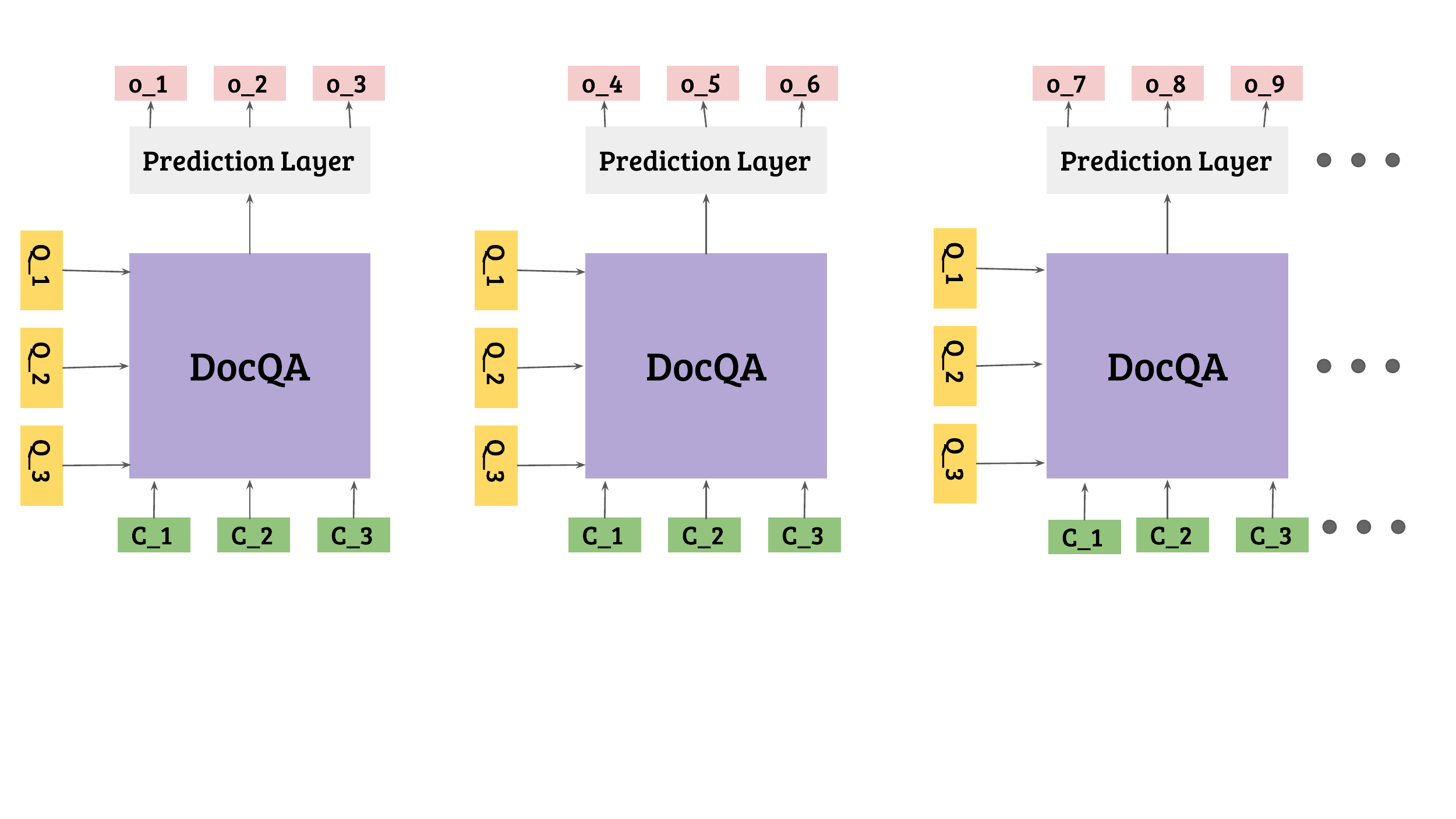}
  \vspace{-40pt}
        \caption{\label{fig:sliced}\fontsize{7}{7}\selectfont{Sliced DocQA}}
    \end{subfigure}%
    ~
\begin{subfigure}[b]{0.4\textwidth}
        \centering
  \includegraphics[width=\textwidth]{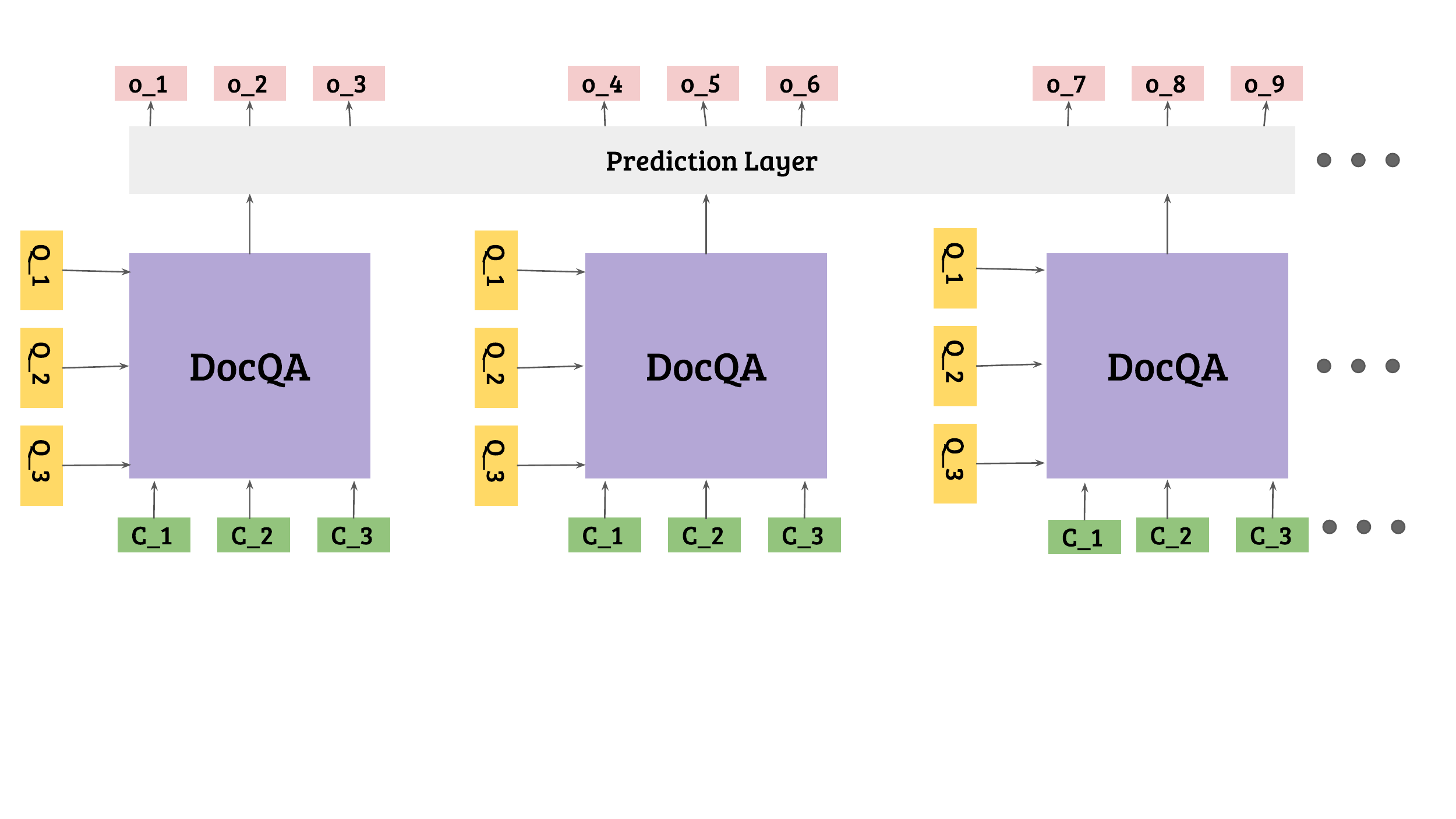}
        \vspace{-40pt}
        \caption{\label{fig:global}\fontsize{7}{7}\selectfont{Global prediction layer}}
    \end{subfigure}
    ~
    \begin{subfigure}[b]{0.4\textwidth}
        \centering
  \includegraphics[width=\textwidth]{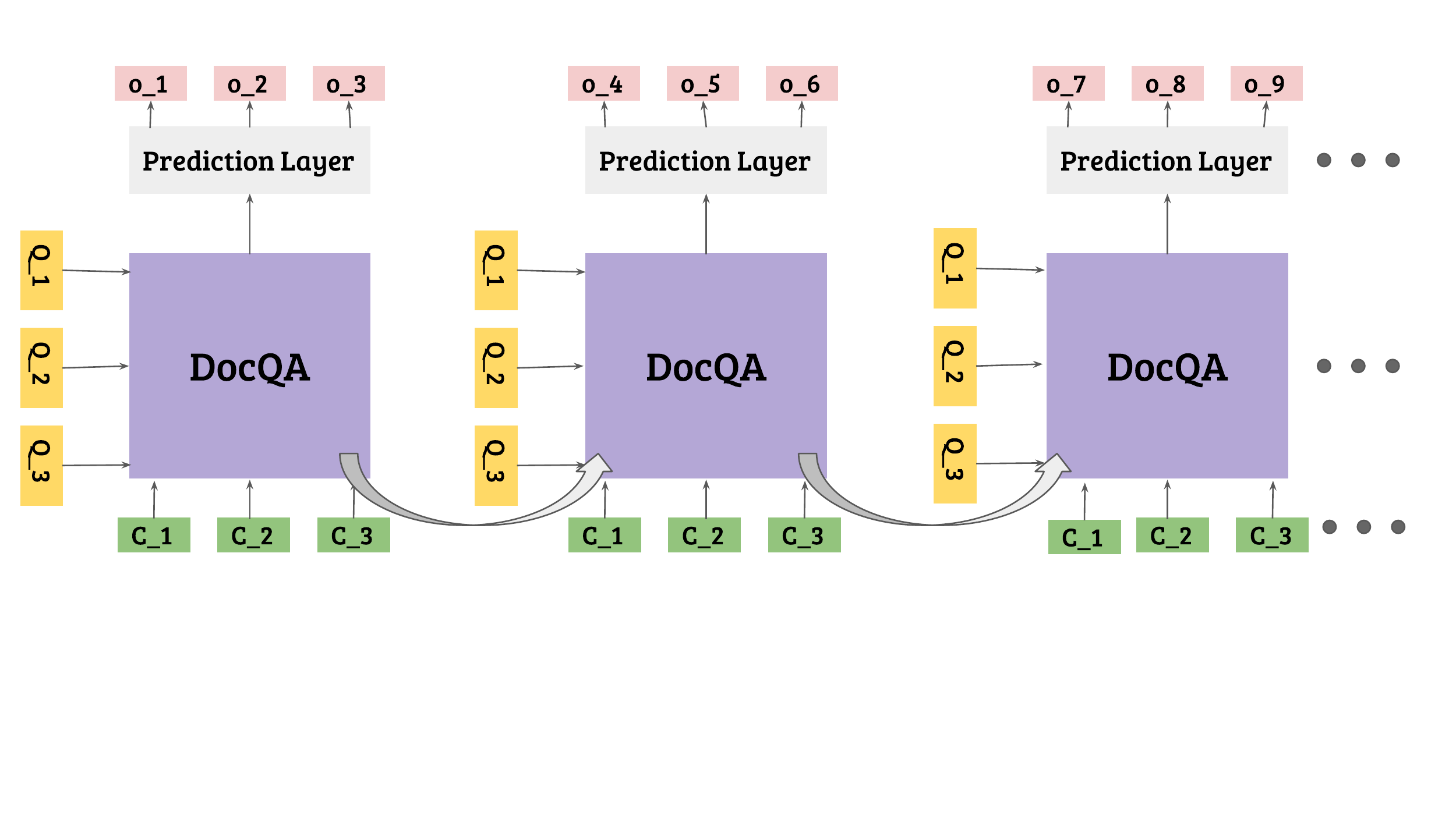}
    \vspace{-40pt}
        \caption{\label{fig:steptransfer}\fontsize{7}{7}\selectfont{Step transfer}}
    \end{subfigure}%
    \caption{\fontsize{8}{8}\selectfont{Sliced Models}}
        \vspace{-10pt}
\end{figure}

\paragraph{Sliced DocQA}
In order to enable the DocQA model to process the context sequence incrementally despite having bidirectional RNN and attention layers, we use the concept of slicing \cite{yu2018sliced}. We divide the context sequence into slices and apply the model to each slice. We call this ``Sliced DocQA''. We explore different ways of using Sliced DocQA for incremental question answering.
\thiswillnotshow{
Then aggregate the outputs for different slices into a final answer. For now, we have two simple strategies for the aggregation part: concatenate the sliced outputs together or apply a global prediction layer.
}
\paragraph{Sliced Prediction Layer}
In the simple sliced model, we predict the output for each slice independently. Thus, as the model processes each slice, it predicts whether the answer span exists in that slice or not.  We call this \emph{sliced prediction} (see Figure \ref{fig:sliced}). To aggregate, we use a  \textit{softmax} layer on the concatenation of the outputs of all slices.
This architecture allows us to processes the input incrementally, but each slice is now processed independently: i.e., when reading the second slice, the model can make no use of what was read in the first slice. This architecture hence ignores the order of the slices. We propose two solutions to solve this issue, discussed in more detail below. The first solution is to let the model access all representations from all slices in the prediction layer: we call this strategy \emph{global answer prediction}. The second solution is a novel mechanism to transfer knowledge between slices called \emph{step transfer}. 

\paragraph{Global Answer Prediction}
While the sliced model processes the context sequence in slices, in the last layer, we use the encoded information from all the slices of the context to predict the answer. This means the prediction layer is not sliced. This model is illustrated in Figure \ref{fig:global}. Global answer prediction can be made incremental by applying the prediction once for each slice, with information from the future slices masked out.
\paragraph{Step Transfer}
To connect slices incrementally, we can transfer the knowledge between slices by using the knowledge learned until the current slice to initialize the next slice, as illustrated in Figure \ref{fig:steptransfer}.
Thus, at the global level we have a uni-directional RNN, but, locally, we can have bidirectional or self-attention layers.
To do this, we use encoded information from the current slice as an input to a fully connected network to predict the initial state for the next slice.
\thiswillnotshow{
In our model, we have step transfer for the contextualization and modelling layers in DocQA.
Input for the step transfer in the contextualization layer is the output of the attention layer averaged over all the positions in the slice.
Input for the step transfer in the modelling layer is the output of the modelling layer averaged over all the positions in the current slice.}

\paragraph{Early Stopping}
These incremental models can be used to support \emph{early stopping} \cite{dulac2011text,yu2017learning}. We use a supervised approach to predict when the model should stop, with a classifier that is simply trained on detecting whether an answer is contained in a given context or not. We train this classifier in a multi-task framework by adding an extra objective function to the QA system. Hence, at each training step, the model not only tries to predict an answer but also predicts whether the true answer is within the currently processed input or not. The early stopping classifier is a two-layer fully connected neural network with \textit{RELU} activations and a \textit{sigmoid}  output layer. The input to this network is the average of the representations in the last layer of the current slice, and the output is a scalar indicating the probability of stopping. At test time the overall decision as to whether to stop early is based on thresholding the cumulative sum of these predictions, i.e. the probability of stopping at slice $i$ is the sum of the predicted probabilities of stopping at slices $1$ to $i$. 
Equation \ref{eq:eloss} shows how we compute the loss for the early stopping model. 
\begin{equation}
  stop\_loss = \sum_{i}{(predicted\_stop\_prob_i - gold\_stop\_label_i)^{2} * extra\_length_i}
\label{eq:eloss}
\end{equation}
Here $extra\_length_i$ is a factor for scaling the early stopping loss, based on the distance from the gold stop point, in particular: $extra\_length = \log(
          \max(dist\_threshold,\abs{length\_read - answer\_end}))$.
In this equation, $dist\_threshold$ is a minimum number of tokens after which we start scaling the loss. Thus, in case the model has not yet reached the answer and decides to stop it will be punished more if it is too early in contrast to when it will reach the answer by reading a few extra tokens. Similarly, if the model has already passed the answer span, the more it reads, the bigger the loss will be. If we are in a distance less than the $dist\_threshold$ from the gold stop point, the scaling factor is $\log(dist\_threshold)$.
In the end, this loss is added to the answer prediction loss to form the total loss.  If the true answer span is not before the chosen stopping point, the answer prediction loss is set to zero for all possible answers, and the model is only trained to choose a better stopping point.


\thiswillnotshow{
The combination of reading the text incrementally and learning to stop when the reading goal is reached can be viewed as planning the reading, where the reading strategy is fixed, and the planning is basically, deciding whether it should stop or not. This problem is addressed for tasks like text classification \cite{dulac2011text} and question answering \cite{shen2017reasonet}. This can be taken one step further, where the model can also skip the irrelevant content by jumping in the text sequence \cite{yu2017learning, johansen2017learning}.
}

\section{Experiments and Discussion}
\usetikzlibrary{patterns}

\pgfplotstableread{
SliceSize Sliced StepTransfer GlobalPrediction 
1	0.31396437	0.59639441	0.74043602
4	0.50739263	0.68621604	0.76343156
8	0.59154002	0.71697662	0.77479835
16	0.66713823	0.74189297	0.7740389
32	0.71743157	0.76153738	0.77866965
64	0.76059337	0.7696418	0.7815346
128	0.78043389	0.78307083	0.76271986
}\tableone

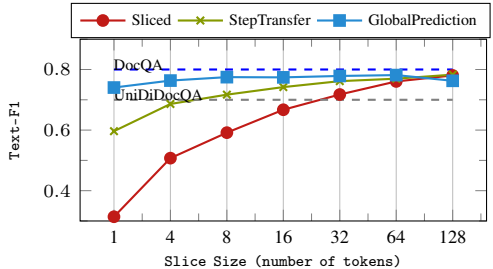
\begin{wrapfigure}{r}{0.5\textwidth}
\centering
\vspace{-40pt}
\begin{tikzpicture}
\begin{axis}[
width= 0.5*\textwidth,
height=4cm,
xmajorgrids,
minor tick num=1,
ylabel={\texttt{Text-F1}},
xlabel={\texttt{Slice Size (number of tokens)}},
xtick=data,
ymin=0.30, 
ymax=0.9,
ticklabel style = {font=\fontsize{7}{8}\selectfont},
xtick={1,4,8,16,32,64,128,512,1024},
xticklabels={1,4,8,16,32,64,128,512,1024},
symbolic x coords={1,4,8,16,32,64,128,512,1024},
label style = {font=\fontsize{6}{7}\selectfont, yshift=0.5ex},
legend style={at={(1.01,1.2),font=\fontsize{6}{7}\selectfont}
  ,legend columns=4
  },
]
\addplot [thick, gcolor, mark=*] table[x index=0, y index=1]{\tableone};

\addplot [thick, wcolor, mark=x] table[x index=0,y index=2] {\tableone};

\addplot [thick, lcolor , mark=square*] table[x index=0,y index=3]
{\tableone};

\addplot+[ thick, gray, dashed,
    no markers,
    decoration={
        text along path,
        text format delimiters={<}{>},
        text={<\tiny>UniDiDocQA},
    },
    postaction={decorate},
] coordinates {(1,0.7) (128,0.7)};

\addplot+[ thick, blue, dashed,
    no markers,
    decoration={
        text along path,
        text format delimiters={<}{>},
        text={<\tiny>DocQA},
    },
    postaction={decorate},
] coordinates {(1,0.8) (128,0.8)};

\legend{Sliced, StepTransfer, GlobalPrediction}
\end{axis}
\end{tikzpicture}
\caption{\label{fig:results}\fontsize{8}{8}\selectfont{Performance of different versions of the Sliced DocQA with respect to different slice sizes}}
\vspace{-25pt}
\end{wrapfigure}

\paragraph{Task and Dataset: \squad}
We study the performance of incremental models on short answer questions answering.

We Evaluate our model on \squad~ v1 \cite{rajpurkar2016squad} which is a reading comprehension dataset. It contains question and context pairs, where the answer to the question is a span in the context. 
\thiswillnotshow{
Table \ref{table:dataset_stats} shows some statistics about the dataset.
\begin{table}[ht]
\resizebox{\textwidth}{!}{%
\begin{tabular}{ccccccccccccccc}
 & \rot{Number of Examples} & \rot{Avg Context Length} & \rot{Min Context Length} & \rot{Max Context Length} & \rot{Avg Min Answer Length} & \rot{Avg Max Answer Length} & \rot{Avg Min Answer Start} & \rot{Avg Min Answer End} & \rot{Avg Max Answer Start} & \rot{Avg Max Answer End} & \rot{Min Answer Start} & \rot{Max Answer End} & \rot{Min Answer Length} & \rot{Max Answer Length} \\\midrule
\squad (train) & 87599 & 141.93 & 22 & 816  & 3.52 & 3.52 & 59.74 & 62.26 & 59.74  & 62.26 &  0 & 634 & 1 & 46  \\ \midrule
\squad (dev) & 10570 & 146.15  & 26 & 708 & 2.42  & 4.16 & 58.68 & 60.81  &  63.85 & 66.24 & 0 & 569 & 1 & 37   \\  \midrule
\bottomrule
\end{tabular}
}
\caption{
\label{table:dataset_stats}
Statistics of \squad~ v1}
\end{table}
}
We will open source the code for reproducing the experiments.
\subsection{Experiments}
We experiment with different slice sizes for Sliced DocQA in three different modes:
with a sliced prediction layer,
with a sliced prediction layer and step transfer,
without step transfer but with a global prediction layer.
In the sliced models, the size of each slice can play a crucial role. The extreme cases are when the length of the slices are 1, which means we have no bidirectional layer, and when the slices are as long as the whole sequence, which means we have a fully bidirectional model. Notice that when slice size is 1, the sliced model with step transfer is not equivalent to the uni-directional DocQA, since the attention layer is implemented differently.
\paragraph{Effect of slice size}
Figure \ref{fig:results} shows the performance of the Sliced DocQA models with respect to different slice sizes. With a sliced prediction layer, as expected, increasing the slice size leads to an increase in the performance. In this case, having a slice size of 1 means to predict the answer based on single word representations and still, we can get an accuracy of $31\%$. We can explain this by the fact that $30\%$ of answers in the \squad~ are single words, e.g. when the questions are of type when, where or who.
\thiswillnotshow{If this does not happen at all, it means our models are not using any information from the context to predict the logits that correspond to each token.} 
We expect to find a slice size for which the performance of the model reaches the performance of the complete, not sliced, model. This is because, only local attention \thiswillnotshow{replace local attention with something better} might be enough to correctly understand the meaning of the input sequence at each step, and also to predict whether each token is part of the answer or not. We observe that by slice size = 32, the model already reaches the performance of the uni-directional DocQA, and by slice size = 128 it almost reaches the performance of the normal DocQA.
The interesting finding here is that there is a limit to which increasing the slice size leads to an increase in the performance, e.g. for the \squad, as soon as the slice size reaches 128 tokens, the increase in the performance is almost not noticeable.

\paragraph{Step transfer is useful}
When we have a step transfer mechanism, we observe that for each slice size, the performance of the model is higher. This means transferring knowledge between slices is useful.
Among these models, the Sliced DocQA with step transfer models incremental comprehension and we see that for a slice size of 64 its performance is comparable with normal DocQA. 

\paragraph{Global answer prediction  is effective}
Surprisingly, with the global prediction layer, the effect of the slice size is much less. 
In general, when we have a global prediction layer, the performance of the model with all slice sizes is higher than when we have a sliced prediction layer. It is interesting to see that in this model, even with slice size of 1, which means using word embeddings to predict the answer span, we still get a performance close to when we have larger slice sizes.

\begin{figure}[!t]
\centering
\begin{subfigure}[b]{0.40\textwidth}
\includegraphics[width=\textwidth]{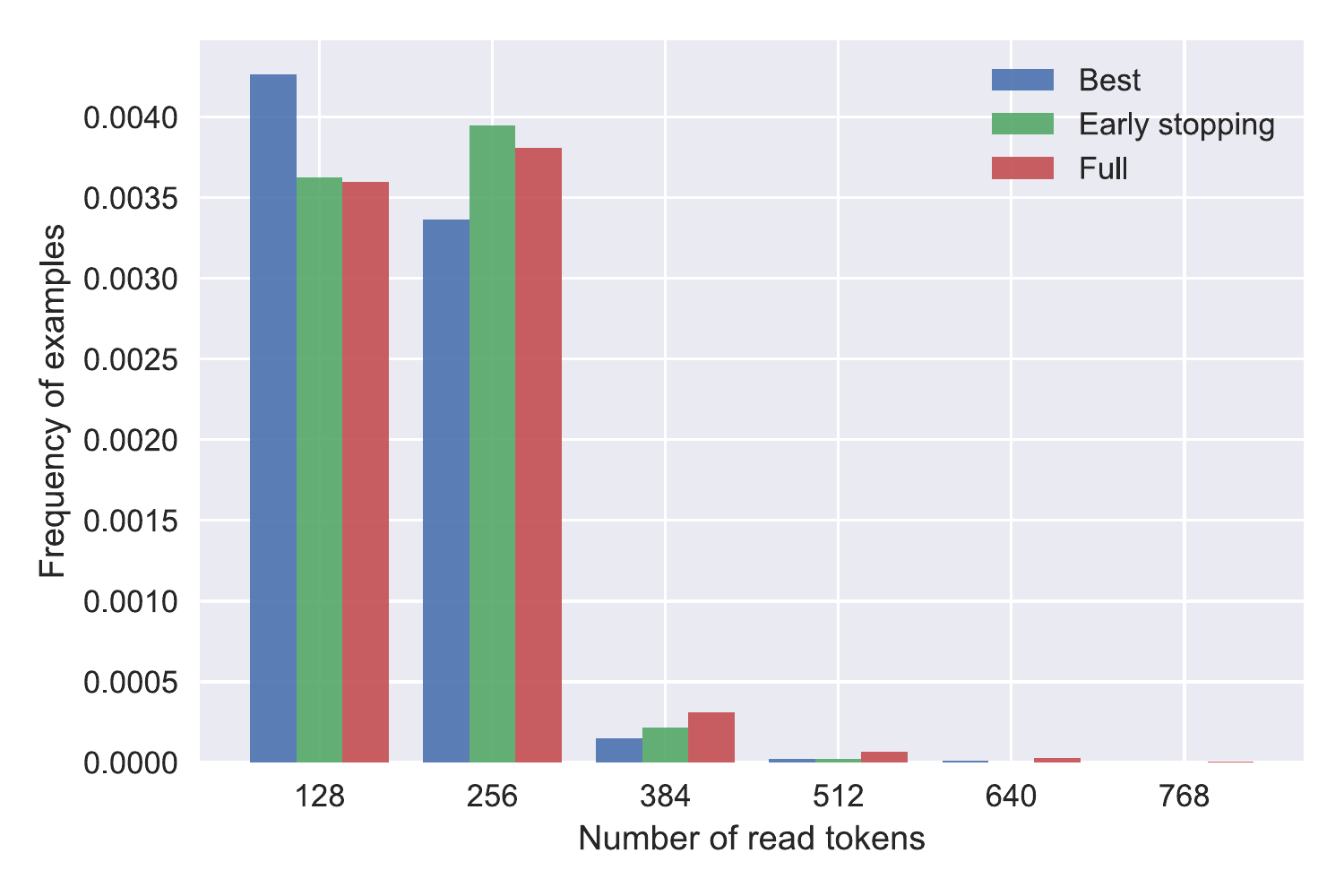}
\caption{\label{fig:es_results}\fontsize{8}{8}\selectfont{Distribution of consumed length of the sequence.}}
\end{subfigure}
~
\begin{subfigure}[b]{0.43\textwidth}
\centering
\includegraphics[width=\textwidth]{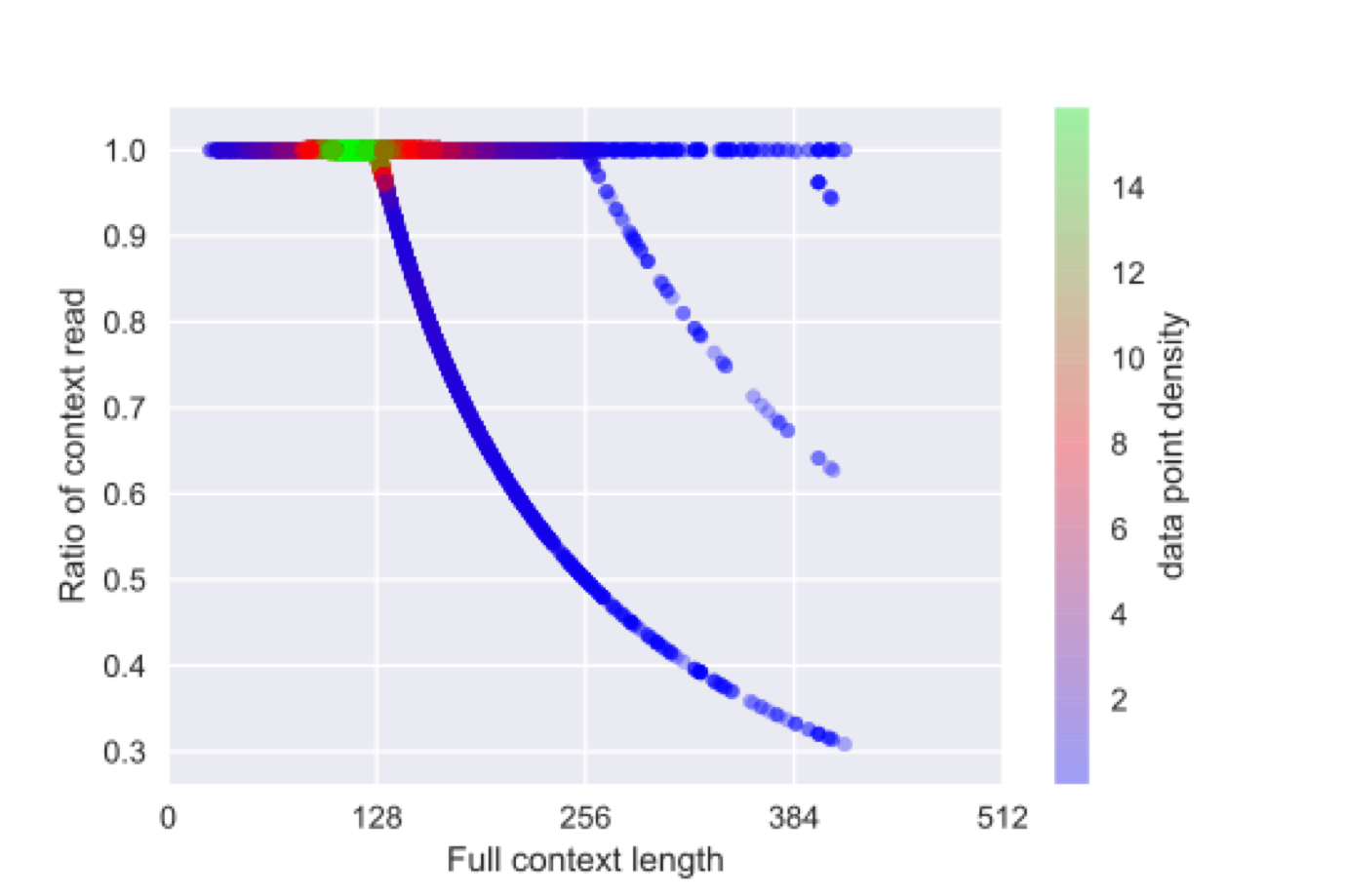}
\caption{\label{fig:earlystopping}\fontsize{7}{8}\selectfont{Ratio of consumed context per full context length for the early stopping model.}}
\end{subfigure}
\caption{\fontsize{8}{8}\selectfont{Earlystopping Results (For the earlystopping model slice size is 128).}}
\vspace{-10pt}
\end{figure}

\begin{table}[!h]
\centering
\fontsize{8}{8}\selectfont{
\begin{tabular}{l|c|c|c}
\textit{Slice size} & 32 & 64 & 128 \\
\midrule
\textit{Sliced DocQA w/o early stopping} & 0.76 & 0.77 & 0.78\\
\midrule
\textit{Sliced DocQA w early stopping}& 0.54 (\%71) & 0.69(\%90) & 0.77 (\%99)\\
\end{tabular}
}
\caption{\fontsize{8}{10}\selectfont{Performance of Sliced DocQA with step transfer and early stopping in terms of text-f1.}}
\label{tab:earlystop_res}
\vspace{-20pt}
\end{table}
\paragraph{Effect of early stopping}
In our early stopping experiments, we employed Sliced DocQA with Step Transfer as a truly incremental model.
In Table~\ref{tab:earlystop_res}, we compare the performance of this model with different slice sizes. As it can be seen, at slice size of 128 it almost achieves the performance of the model without early stopping ($\%99$). 
Next, we investigate whether the early stopping model is more efficient regarding the number of read tokens.
We looked into the performance of the model without early stopping at different context length to find the earliest point at which the model achieves its highest performance. This can be assumed as an oracle for early stopping model (i.e. best possible stopping point). In Figure \ref{fig:es_results} we compare the distribution of best stopping points with the stopping points of our early stopping model. We observe that (1)For a large number of examples, we have to read the full context to achieve the best performance. (2)For some examples, the early stopping model is reading more than it should, and for some, it stops earlier than it should. (3)In general, while with the best stopping offsets we can read about $\%15$ less text, our model reads about $\%8$ less. 
We also studied if there is a correlation between the context length and the ratio of read context length. In Figure \ref{fig:earlystopping}, we see for context length up to 128, we read full context, which is because our slice size is 128. After that point, we see the trend of reading smaller ratio of longer contexts.
\section*{Conclusion}
\vspace{-5pt}
In this paper, we propose a model that reads and comprehends text incrementally. As a testbed for our approach, we have chosen the question answering task. We aim to build a model that can learn incrementally from text, where the learning goal is to answer a given question. In standard question answering, we do not care how the context is presented to the model, and for the models that achieve state of the art results, e.g. \cite{yu2018qanet, clark2017simple}, they process the full context before making any decisions. We show that it is possible to modify these models to be incremental while achieving similar performance. Having an incremental model, allows us to employ an early stopping strategy where the model avoids reading the rest of the text as soon as it reaches a state where it thinks it has the answer.

\bibliography{refs} 

\begin{thebibliography}{10}

\bibitem{battaglia2018relational}
Peter~W Battaglia, Jessica~B Hamrick, Victor Bapst, Alvaro Sanchez-Gonzalez,
  Vinicius Zambaldi, Mateusz Malinowski, Andrea Tacchetti, David Raposo, Adam
  Santoro, Ryan Faulkner, et~al.
\newblock Relational inductive biases, deep learning, and graph networks.
\newblock {\em arXiv preprint arXiv:1806.01261}, 2018.

\bibitem{clark2017simple}
Christopher Clark and Matt Gardner.
\newblock Simple and effective multi-paragraph reading comprehension.
\newblock In {\em Proceedings of the 56th Annual Meeting of the Association for
  Computational Linguistics (Volume 1: Long Papers)}, pages 845--855.
  Association for Computational Linguistics, 2018.

\bibitem{dulac2011text}
Gabriel Dulac-Arnold, Ludovic Denoyer, and Patrick Gallinari.
\newblock Text classification: A sequential reading approach.
\newblock In {\em European Conference on Information Retrieval}, pages
  411--423. Springer, 2011.

\bibitem{hermann2015teaching}
Karl~Moritz Hermann, Tomas Kocisky, Edward Grefenstette, Lasse Espeholt, Will
  Kay, Mustafa Suleyman, and Phil Blunsom.
\newblock Teaching machines to read and comprehend.
\newblock In {\em Advances in Neural Information Processing Systems}, pages
  1693--1701, 2015.

\bibitem{hochreiter1997long}
Sepp Hochreiter and J{\"u}rgen Schmidhuber.
\newblock Long short-term memory.
\newblock {\em Neural computation}, 9(8):1735--1780, 1997.

\bibitem{johansen2017learning}
Alexander~Rosenberg Johansen and Richard Socher.
\newblock Learning when to skim and when to read.
\newblock {\em arXiv preprint arXiv:1712.05483}, 2017.

\bibitem{rajpurkar2016squad}
Pranav Rajpurkar, Jian Zhang, Konstantin Lopyrev, and Percy Liang.
\newblock Squad: 100,000+ questions for machine comprehension of text.
\newblock In {\em Proceedings of the 2016 Conference on Empirical Methods in
  Natural Language Processing}, pages 2383--2392. Association for Computational
  Linguistics, 2016.

\bibitem{seo2016bidirectional}
Minjoon Seo, Aniruddha Kembhavi, Ali Farhadi, and Hannaneh Hajishirzi.
\newblock Bidirectional attention flow for machine comprehension.
\newblock {\em arXiv preprint arXiv:1611.01603}, 2016.

\bibitem{shen2017reasonet}
Yelong Shen, Po-Sen Huang, Jianfeng Gao, and Weizhu Chen.
\newblock Reasonet: Learning to stop reading in machine comprehension.
\newblock In {\em Proceedings of the 23rd ACM SIGKDD International Conference
  on Knowledge Discovery and Data Mining}, pages 1047--1055. ACM, 2017.

\bibitem{traxler1997influence}
Matthew~J Traxler, Michael~D Bybee, and Martin~J Pickering.
\newblock Influence of connectives on language comprehension: eye tracking
  evidence for incremental interpretation.
\newblock {\em The Quarterly Journal of Experimental Psychology: Section A},
  50(3):481--497, 1997.

\bibitem{yu2018qanet}
Adams~Wei Yu, David Dohan, Minh-Thang Luong, Rui Zhao, Kai Chen, Mohammad
  Norouzi, and Quoc~V Le.
\newblock Qanet: Combining local convolution with global self-attention for
  reading comprehension.
\newblock {\em arXiv preprint arXiv:1804.09541}, 2018.

\bibitem{yu2017learning}
Adams~Wei Yu, Hongrae Lee, and Quoc Le.
\newblock Learning to skim text.
\newblock In {\em Proceedings of the 55th Annual Meeting of the Association for
  Computational Linguistics (Volume 1: Long Papers)}, pages 1880--1890.
  Association for Computational Linguistics, 2017.

\bibitem{yu2018sliced}
Zeping Yu and Gongshen Liu.
\newblock Sliced recurrent neural networks.
\newblock In {\em COLING}, 2018.

\end{thebibliography}
\bibliographystyle{plain}

\medskip

\small

\newpage
\section*{Appendix}

\subsection{Architecture of Baseline Models}

\end{document}